\documentclass[conference]{IEEEtran}
\IEEEoverridecommandlockouts

\usepackage{booktabs, multirow} 
\usepackage{soul}
\usepackage[table]{xcolor} 
\usepackage{changepage,threeparttable} 

\usepackage{cite}
\usepackage{amsmath,amssymb,amsfonts}
\usepackage{algorithmic}
\usepackage{graphicx}
\usepackage{textcomp}
\usepackage{xcolor}

\usepackage{tikz}
\usetikzlibrary{positioning, arrows.meta}

\def\BibTeX{{\rm B\kern-.05em{\sc i\kern-.025em b}\kern-.08em
    T\kern-.1667em\lower.7ex\hbox{E}\kern-.125emX}}

\begin{document}
\title{Tag Prediction of Competitive Programming Problems using Deep Learning Techniques}

\makeatletter
\newcommand{\linebreakand}{
  \end{@IEEEauthorhalign}
  \hfill\mbox{}\\ 
  \mbox{}\hfill\begin{@IEEEauthorhalign}
}
\makeatother

\author{
    \IEEEauthorblockN{Taha Lokat}
    \IEEEauthorblockA{\textit{Department of Computer Engineering} \\
    \textit{K. J. Somaiya Institute of Technology}\\
    Mumbai, India\\
    taha.lokat@somaiya.edu}
    \and
    \IEEEauthorblockN{Divyam Prajapati}
    \IEEEauthorblockA{\textit{Department of Computer Engineering} \\
    \textit{K. J. Somaiya Institute of Technology}\\
    Mumbai, India\\
    divyam.prajapati@somaiya.edu}
    \and
    \IEEEauthorblockN{Shubhada Labde}
    \IEEEauthorblockA{\textit{Department of Computer Engineering} \\
    \textit{K. J. Somaiya Institute of Technology}\\
    Mumbai, India\\
    shubhada.l@somaiya.edu}
}

\maketitle
\thispagestyle{plain}
\pagestyle{plain}
\begin{abstract}
In the past decade, the amount of research being done in the fields of machine learning and deep learning, predominantly in the area of natural language processing (NLP), has risen dramatically. A well-liked method for developing programming abilities like logic building and problem solving is competitive programming. It can be tough for novices and even veteran programmers to traverse the wide collection of questions due to the massive number of accessible questions and the variety of themes, levels of difficulty, and questions offered. In order to help programmers find questions that are appropriate for their knowledge and interests, there is a need for an automated method. This can be done using automated tagging of the questions using Text Classification. Text classification is one of the important tasks widely researched in the field of Natural Language Processing. In this paper, we present a way to use text classification techniques to determine the domain of a competitive programming problem. \textit A variety of models, including are implemented LSTM, GRU, and MLP. The dataset has been scraped from codeforces, a major competitive programming website. A total of 2400 problems were scraped and preprocessed, which we used as a dataset for our training and testing of models. The maximum accuracy reached using our model is 78.0\% by MLP(Multi Layer Perceptron). \\ 
\end{abstract}


\section{Introduction}
    A tremendous amount of research has been done in the past decade in the area of Natural Language Processing (NLP), which is concerned with how computers and human languages interact. There are many different applications of NLP, such as Sentiment Analysis, Text Classification, Machine Translation, text summarization, and many more~\cite{wankhade2022survey, kowsari2019text, jiang2020natural}. Text classification is an application of NLP that involves categorizing or labeling a given text based on its content. With many applications in areas including sentiment analysis, topic modeling, spam detection, and more, text categorization is a fundamental NLP issue. In general, text categorization may be divided into two categories binary and multi-class. In binary classification, the text is classified into one of the two predetermined groups, whereas in multi-class classification, it is classified into one of the many specified groups. On the other hand, multi-label classification is a subset of multi-class classification in which a single text may simultaneously belong to several categories.\par  

    Multi-class classification is a difficult issue in NLP that has attracted a lot of interest lately. Each text instance in this job can be connected to many labels. Multi-class classification is prevalent in various real-world applications, such as document categorization, image tagging, and music genre classification. \par    
    
    As there have been advancements in the area of Deep Learning (DL), it has also been used for solving these NLP problems. Convolutional neural networks (CNNs) and Recurrent Neural Networks (RNNs) are two of the most commonly used deep learning approaches that have been successfully used in multi-class classification. ~\cite{liu2017deep}. This paper also provides a DL based approach that can be used for multi-label text classification purposes. This paper is organized in the following manner: Section I contained the introduction to multi-class text classification; Section II contained the Literature survey; and Section III contained our methodology, i.e., the proposed model along with details about the dataset used and the pre-processing techniques applied to it, as well as training and testing techniques. Section IV is Results and Discussion, which discusses how our approach is better than previous approaches and comparisons based on metrics. Lastly, Section V concludes the paper and also provides some future direction. \par  
    
\section{Related Work}
    There has been numerous amount of research that has been on the topic of multi-label text classification especially using DL techniques in the past decade and we will go thorough some of the literature in this section. \par

    
    The comparision between 12 machine learning pipelines using the dataset Enron spam corpus is done~\cite{occhipinti2022pipeline}. The preprocessing steps involve removal of stop words, lemmatisation, removal of HTML tags and single letters and numbers. After preprocessing using the above steps, various machine learning algorithms were fit on the processed data. The machine learning algorithms used are Naive Bayes, Support Vector Machine (SVM), k-Nearest Neighbours(kNN), Multi Layer perceptron Neural Network(MLP), Logistic Regression, Random Forest and Extreme Gradient Boost (XGBoost). The authors use 5-fold cross validation throughout the test cases to avoid overfitting the model. Random Forest was the best performing model with precision, recall and F1-score of 0.94, 0.94, 0.4 respectively.

    Several large datasets like AG's news corpus, Sogou news corpus, DBPedia onotology dataset, Yelp reviews, Yahoo! answers dataset and the Amazon reviews dataset to present a character level ConvNet~\cite{zhang2015character}. There were two ConvNets designed : one large and one small each of them nine layers deep with six convolutional layers and three fully connected layers The most important conclusion from the experiments was that character-level ConvNets could work for text classification without the need for words. This meant there was a strong indication that language can also be thought of as a signal no different from any other kind.

    An improved class specific word vector taht enhances the distinctive property of word in a class to tackle light polysemy problem in question classification~\cite{gupta2021learning}. The models used are Convolutional Neural Networks(CNN), Bidirectional LSTM(Bi-LSTM) and Attention Based Bi-GRU CNN(ABBC). The accuracy on TREC dataset was 0.936, the accuracy on Kaggle questions dataset was 0.918 and the accuracy on Yahoo questions dataset was 0.892. \par 

    The methods to improve text classfication on large number of unbalanced classes and noisy data are given~\cite{fonseca2020automatic}. The dataset used contains 57,647 English song texts with their artist and title that is downloaded from kaggle. The models used are perceptron, Doc2vec and Multilayer Perceptron(MLP). There are two versions of Perceptron used : the minimal version(Perceptron) and maximal version (Perceptron+). The tokenization for the Doc2Vec model is performed using the UIMA tokenizer. There are also two versions of MLP: MLP and MLP+ that has one bias feature for every group. MLP+ performs the best with F-score of 0.079 on training set and 0.182 on test set. The worst performing model is Perceptron+ with F-score of 0.003 on train set and 0.021 on 
    the test set. \par

    Many different types of EDA(Easy Data Augmentation) techniques are given which include Synonym Replacement(SR), Random Insertion(RI), Random Swap(RS) and Random Deletion(RD)~\cite{wei-zou-2019-eda}. RNNs (Recurrent Neural Networks) and CNNs (Convolutional Neural Networks) are then ru on five NLP tasks with and without the aforementioned EDA steps and an improvement was observed on the full dataset.

    A Universal Language Model Fine tuning (ULMFiT) is proposed which when given a large corpus of a particular domain then fine tunes an already existing Language Model (LM)~\cite{howard-ruder-2018-universal}. This method is tested on six widely studied datasets used in three most common text classification tasks: Sentiment Analysis, Question Classification and Topic Classification. 

    A text classifier called fastText is proposed which is a simple and efficient baseline model for text classification. It takes as input normalized bag of features of the Nth document (N being the number of documents) and then passes them through a hidden layer with hierarchical softmax~\cite{joulin-etal-2017-bag}. For tag prediction, YFCC100M dataset is used, It was observed that adding of bigrams to the hidden layers improved the accuracy.

    BertGCN is proposed, it uses BERT representations and converts into a heterogeneous graph over the dataset~\cite{lin-etal-2021-bertgcn}. The input representations for document nodes to the BertGCN model are the document embeddings that are obtained using the BERT-style model. The output is then passed to a softmax layer for classification. The model is optimized using a memory bank M that keeps a track of the input features of all the document nodes. 
    
    A BAE is proposed which leverages the BERT-MLM to generate alternative of the masked tokens in the document~\cite{garg-ramakrishnan-2020-bae}. It also replaces a token in the document with another token some of which contribute towards the final prediction.
    
    Convolutional Neural Networks (CNNs) capture the bias caused by keywords appearing everywhere in the text, not only towards the end. A recurrent structure which is a bidirectional recurrent neural network is used in the proposed model to capture the contexts~\cite{Lai_Xu_Liu_Zhao_2015}. The word embedding in the model are pretrained using the Skip-gram model. The accuracy achieved by the model when using Convolutional Neural Network (CNN) is 94.79 while the accuracy achieved by using Recurrent Convolutional Neural Networks (RCNN) is 96.49.

    A Graph neural Network is a multi layer neural network that operates directly on graphs and properties of the neighbourhoods of the nodes are used to induce embeddings of vectors of nodes~\cite{kipf2016semi}. The proposed model TextGCN builds a heterogeneous graph which models global word concurrence explicitly by taking into account word nodes and document nodes~\cite{Yao_Mao_Luo_2019}. The accuracy of TextGCN model is better than most of the models used like CNN, LSTM, etc. The maximum accuracy achieved was 0.9797 with a range of 0.0010 on the R8 dataset.

\section{Methodology}
\subsection{Dataset Collection}

    We have gathered our data for the dataset creation from: Codeforces\footnote{https://codeforces.com/}. Codeforces is a well-known website used by many programmers to increase their logic building, debugging, problem solving, and other programming skills. It has problems in the form of contests, and one contest has many problems depending on the type of contest. Also, the level of problems keeps increasing as we move forward in the contest. So for collecting the data of these contest from these site we made use of a python library called BeautifulSoup\footnote{https://beautiful-soup-4.readthedocs.io/en/latest/}. BeautifulSoup is a web scrapper that is used by many for extracting data from HTML or XML files. \par

    \begin{figure}[!htbp]
        \centerline{\includegraphics[width=7cm]{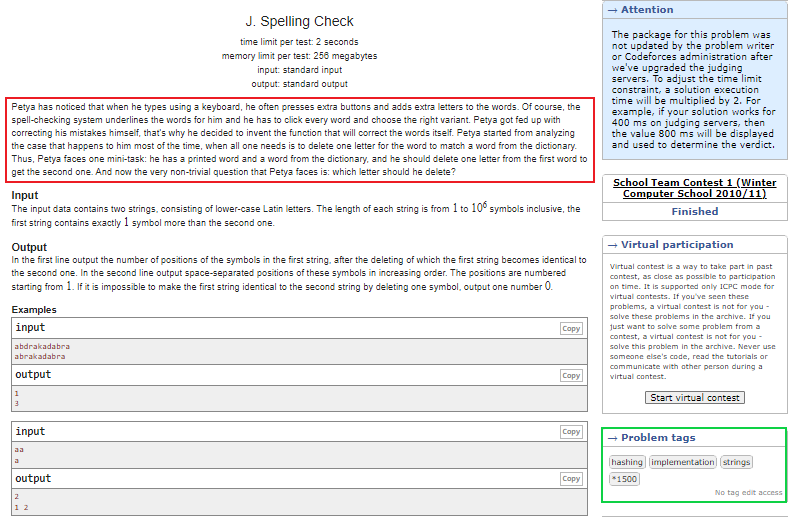}}
        \caption{Problem Page of Codeforces website.}
        \label{fig:ss}
    \end{figure}
        
    The structure of any problem in a contest is shown in Figure No. \ref{fig:ss}, As we can see, there is a main problem statement (highlighted using a red box) that states what the problem is about and what the programmer needs to do in order to solve that problem. Next, there are constraints, which describe the constraints (i.e., the maximum and minimum limits) for the variables mentioned in the problem statement. The input and output sections describe in what format the program will get its input and output, respectively. Next, there will be some sample test cases that will be shown as examples so that the contestants get a clear idea of how their program will work. Lastly, there will be some tags (highlighted using a green box) that are assigned to that problem, which give hints to the solution of that problem. So for multi-label text classification, we will need the problem statements and the tags assigned to that question. First, we scraped its contest page and gathered all the problem ids for each contest, and then for each problem id, we scraped its problem statement and tags assigned to it, and lastly, we stored it in a Pandas dataframe and exported it in the form of a CSV. Now that our two dataset was ready, we started to preprocess it, which is covered in the next subsection. \par

\subsection{Pre-processing Data}

    As soon as the CSV was ready, we started to pre-process it, for which the first and most important step was to decide which tags we wanted to consider in our dataset. For pre-processing tags, all the special characters were removed from the tag name, and all the tags were scraped in the form of a string, which was converted into a list, and only those 3 tags were kept; the rest were deleted, and now all the rows that were empty (nan) were removed from the dataset, so that we can get only the problem statements that have those 3 tags in them. After all the unwanted tags were removed, we started to clean up the problem statements, as they had some LaTeX tags and some other unwanted items in them. \par
    
    So first all these unwanted characters were removed  and secondly all the sentences were converted into lowercase then tokenized using ``word\_tokenize'' from NLTK\footnote{https://www.nltk.org/} library. After which all the stop words were removed. Stop words are those words that are filtered out in the process of NLP (like prepositions, conjunctions, pronouns, articles, etc.). Once the stop words were removed, the text was lemmitized using ``WordNetLemmatizer'' from NLTK. Once all the cleaning and preprocessing was done, both datasets were combined, and finally we got our dataset, which had 992 problems and 3 tags in it. The 3 tags selected were greedy, graphs and implementation. If a problem had multiple only the tag belonging to the aforementioned 3 tags was selected and the others were discarded. 
    
    After the generation of the dataset, the problem statements in the generated dataframe were broken down into individual words using the word tokenizer present in the NLTK library. After that, we find out the maximum length of the sequences, which was 951. We then use the Tensorflow tokenizer to split the given problem statements into tokens. This is then used to pad the problem statements to the maximum length. \par
    \begin{figure}[!htbp]
        \centerline{\includegraphics[width=7cm]{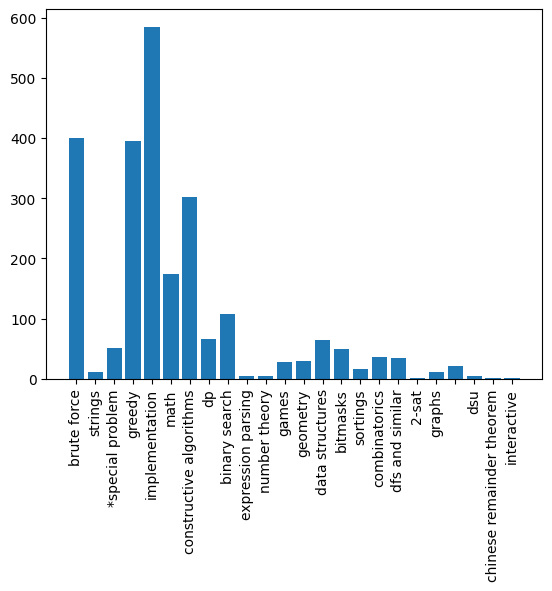}}
        \caption{Codeforces problem statements belonging to different categories by considering one category per problem statement
        }
        \label{fig:chart}
    \end{figure}

\subsection{Proposed Approach}
\subsubsection{Long Short-Term Memory (LSTM) Networks}
    LSTM is a popular approach used in deep learning. Long Short Term Memory (LSTM) is a special kind of Recurrent Neural network capable of learning long-term dependencies~\cite{10.1162/neco.1997.9.8.1735}. They are specifically designed to solve the vanishing gradient problem that comes with a traditional Simple Recurrent Neural Network by remembering information for long periods of time. We initially have the embedding layer with the input dimension as the vocabulary size Instead of having a simple single layer, we have three layers interacting with each other. We one LSTM layer with 32 neurons followed by a dense layer of 32 neurons. The activation function used is ReLU for all layers and Sigmoid for the final output layer. We used the rmsprop optimizer and used categorical cross entropy as the loss function. \par

\subsubsection{Gated Recurrent Units}
    GRU (Gated Recurrent Units): Gated Recurrent Units (GRU) were introduced in 2014 with the aim of solving the vanishing gradient problem that comes with a standard Recurrent Neural Network. GRU has an update gate and a reset gate ~\cite{cho2014learning}. These two gates decide what information should be passed to the output. This helps to keep the information from long ago without washing it through time or removing information that is irrelevant to the prediction, thereby solving the vanishing gradient problem. We have a GRU layers with 32 neurons followed by dense layer of 32 neurons. The final output layer has a Sigmoid activation function, while the ReLU activation function is used for all other layers. Categorical cross entropy is used as the loss function in this case, and rmsprop is the optimizer used. \par
    
\subsubsection{Multi Layer Perceptron}
    A Multi-Layer Perceptron is a feed-forward neural network. In the proposed system of the Multi-Layer Perceptron, we have one input layer, which is a Dense layer with 64 neurons. After that, we add a Dropout layer with a fraction of 0.5. This is followed by another Dense layer of 32 neurons and a Dropout layer with a fraction of 0.5. The final layer is an output layer with 8 neurons. The activation function is used for all layers in ReLu (Rectified Linear Unit), while the final layer has the Sigmoid activation function. We used the rmsprop optimizer and categorical cross entropy as the loss function. \par

\subsection{Training \& Testing of proposed model}
     Ater preprocessing the data using the above methods, we split the traning dataset into training and testing dataset. Because the dataset is small, we split the 992 problems as follows: 950 problems were used for training the models and the remaining 42 problems were used for testing. The input to the model was the input sequence of the problem statement generated by the tokenizer and the output contained a softmax activation which outputs 3 probabilities of each tag. \par
     
\section{Results and Discussions}

     Table~\ref{results} shows the accuracy of the models tried in the paper. It was found that Multi Layer Perceptron (MLP) gave the maximum accuracy.
     The other models used like Long Short Term Memory(LSTM) and Gated Recurrent Units (GRU) did not give a good enough accuracy. The maxmum accuracy that was achieved on the training set for Multi Layer Perceptron(MLP) was 73\% while the training accuracy on Long Short Term Memory(LSTM) was 59\% and the training accuracy on Gated Recurrent Unit(GRU) was 59\%. It can be observed that LSTM and GRU give almost the same accuracy for the problem statement defined. A reason for the poor performance of LSTM and GRU is the lack of training examples and the bias  in the dataset because of which the models are not able to capture the feature vectors effectively. A way to improve that would be to use other competitive programming websites like topcoder (https://topcoder.com) and hackerrank (https://hackerrank.com) to increase the size of our training dataset which will subsequently improve the performance of the models.

    \renewcommand{\arraystretch}{1.6} 
    \setlength{\tabcolsep}{10pt}
    \begin{table}[htbp]
    \centering
    \caption{Models and their accuracy}
    \label{results}
    \begin{tabular}{ p{2in} p{1in} }
    \toprule
    \multicolumn{1}{p{2in}}{{\textbf{Model}}} &
    \multicolumn{1}{p{0.5in}}{{\textbf{Accuracy}}} \\ 
    \midrule
    \multicolumn{1}{p{2in}}{{Multi layer feed forward perceptron neural network}} &
    \multicolumn{1}{p{0.5in}}{{72.00\%}} \\
    \multicolumn{1}{p{2in}}{{Stacked Gated Recurrent Units (GRU)}} &
    \multicolumn{1}{p{0.5in}}{{50.00\%}} \\
    \multicolumn{1}{p{2in}}{{Stacked Long Short Term Memory Cells (LSTM)}} &
    \multicolumn{1}{p{0.5in}}{{50.00\%}} \\
    \bottomrule
    \end{tabular}
    \end{table} 

\section{Conclusion And Future Work}
    Hence, in this paper we present ways to classify competitive programming problems into their subsequent categories. Several widely used deep learning techniques were used. The MLP model was the best performing model giving an accuracy of 72\%. In the future, the methods proposed can be used to classify all types of competitive programming problems on various platforms available. Models can further be refined and fine tuned to classify a greater number of tags Even though we classify the problem statements having only 3 tags, the models show considerable performance and can be improved to improve the accuracy. The scope can further be widened by including multi-label classification which will classify problems not into one category but in multiple categories.\par

\bibliographystyle{ieeetr}
\bibliography{bibliography.bib}

\begin{thebibliography}{10}

\bibitem{wankhade2022survey}
M.~Wankhade, A.~C.~S. Rao, and C.~Kulkarni, ``A survey on sentiment analysis
  methods, applications, and challenges,'' {\em Artificial Intelligence
  Review}, vol.~55, no.~7, pp.~5731--5780, 2022.

\bibitem{kowsari2019text}
K.~Kowsari, K.~Jafari~Meimandi, M.~Heidarysafa, S.~Mendu, L.~Barnes, and
  D.~Brown, ``Text classification algorithms: A survey,'' {\em Information},
  vol.~10, no.~4, p.~150, 2019.

\bibitem{jiang2020natural}
K.~Jiang and X.~Lu, ``Natural language processing and its applications in
  machine translation: A diachronic review,'' in {\em 2020 IEEE 3rd
  International Conference of Safe Production and Informatization (IICSPI)},
  pp.~210--214, IEEE, 2020.

\bibitem{liu2017deep}
J.~Liu, W.-C. Chang, Y.~Wu, and Y.~Yang, ``Deep learning for extreme
  multi-label text classification,'' in {\em Proceedings of the 40th
  international ACM SIGIR conference on research and development in information
  retrieval}, pp.~115--124, 2017.

\bibitem{occhipinti2022pipeline}
A.~Occhipinti, L.~Rogers, and C.~Angione, ``A pipeline and comparative study of
  12 machine learning models for text classification,'' {\em Expert Systems
  with Applications}, vol.~201, p.~117193, 2022.

\bibitem{zhang2015character}
X.~Zhang, J.~Zhao, and Y.~LeCun, ``Character-level convolutional networks for
  text classification,'' {\em Advances in neural information processing
  systems}, vol.~28, 2015.

\bibitem{gupta2021learning}
T.~Gupta and E.~Kumar, ``Learning improved class vector for multi-class
  question type classification,'' in {\em 3rd International Conference on
  Integrated Intelligent Computing Communication \& Security (ICIIC 2021)},
  pp.~113--121, Atlantis Press, 2021.

\bibitem{fonseca2020automatic}
S.~C. Fonseca, F.~D. Pereira, E.~H. Oliveira, D.~B. Oliveira, L.~S. Carvalho,
  and A.~I. Cristea, ``Automatic subject-based contextualisation of programming
  assignment lists.,'' {\em International Educational Data Mining Society},
  2020.

\bibitem{wei-zou-2019-eda}
J.~Wei and K.~Zou, ``{EDA}: Easy data augmentation techniques for boosting
  performance on text classification tasks,'' in {\em Proceedings of the 2019
  Conference on Empirical Methods in Natural Language Processing and the 9th
  International Joint Conference on Natural Language Processing
  (EMNLP-IJCNLP)}, (Hong Kong, China), pp.~6382--6388, Association for
  Computational Linguistics, Nov. 2019.

\bibitem{howard-ruder-2018-universal}
J.~Howard and S.~Ruder, ``Universal language model fine-tuning for text
  classification,'' in {\em Proceedings of the 56th Annual Meeting of the
  Association for Computational Linguistics (Volume 1: Long Papers)},
  (Melbourne, Australia), pp.~328--339, Association for Computational
  Linguistics, July 2018.

\bibitem{joulin-etal-2017-bag}
A.~Joulin, E.~Grave, P.~Bojanowski, and T.~Mikolov, ``Bag of tricks for
  efficient text classification,'' in {\em Proceedings of the 15th Conference
  of the {E}uropean Chapter of the Association for Computational Linguistics:
  Volume 2, Short Papers}, (Valencia, Spain), pp.~427--431, Association for
  Computational Linguistics, Apr. 2017.

\bibitem{lin-etal-2021-bertgcn}
Y.~Lin, Y.~Meng, X.~Sun, Q.~Han, K.~Kuang, J.~Li, and F.~Wu, ``{B}ert{GCN}:
  Transductive text classification by combining {GNN} and {BERT},'' in {\em
  Findings of the Association for Computational Linguistics: ACL-IJCNLP 2021},
  (Online), pp.~1456--1462, Association for Computational Linguistics, Aug.
  2021.

\bibitem{garg-ramakrishnan-2020-bae}
S.~Garg and G.~Ramakrishnan, ``{BAE}: {BERT}-based adversarial examples for
  text classification,'' in {\em Proceedings of the 2020 Conference on
  Empirical Methods in Natural Language Processing (EMNLP)}, (Online),
  pp.~6174--6181, Association for Computational Linguistics, Nov. 2020.

\bibitem{Lai_Xu_Liu_Zhao_2015}
S.~Lai, L.~Xu, K.~Liu, and J.~Zhao, ``Recurrent convolutional neural networks
  for text classification,'' {\em Proceedings of the AAAI Conference on
  Artificial Intelligence}, vol.~29, Feb. 2015.

\bibitem{kipf2016semi}
T.~N. Kipf and M.~Welling, ``Semi-supervised classification with graph
  convolutional networks,'' {\em arXiv preprint arXiv:1609.02907}, 2016.

\bibitem{Yao_Mao_Luo_2019}
L.~Yao, C.~Mao, and Y.~Luo, ``Graph convolutional networks for text
  classification,'' {\em Proceedings of the AAAI Conference on Artificial
  Intelligence}, vol.~33, pp.~7370--7377, Jul. 2019.

\bibitem{10.1162/neco.1997.9.8.1735}
S.~Hochreiter and J.~Schmidhuber, ``Long short-term memory,'' {\em Neural
  Comput.}, vol.~9, p.~1735–1780, nov 1997.

\bibitem{cho2014learning}
K.~Cho, B.~Van~Merri{\"e}nboer, C.~Gulcehre, D.~Bahdanau, F.~Bougares,
  H.~Schwenk, and Y.~Bengio, ``Learning phrase representations using rnn
  encoder-decoder for statistical machine translation,'' {\em arXiv preprint
  arXiv:1406.1078}, 2014.

\end{thebibliography}

\end{document}